# Regularizers versus Losses for Nonlinear Dimensionality Reduction

**Yaoliang Yu, James Neufeld, Ryan Kiros, Xinhua Zhang, Dale Schuurmans**
Department of Computing Science, University of Alberta, Edmonton, AB T6G 2E8 Canada

## Abstract

We demonstrate that almost all non-parametric dimensionality reduction methods can be expressed by a simple procedure: regularized loss minimization plus singular value truncation. By distinguishing the role of the loss and regularizer in such a process, we recover a factored perspective that reveals some gaps in the current literature. Beyond identifying a useful new loss for manifold unfolding, a key contribution is to derive new convex regularizers that combine distance maximization with rank reduction. These regularizers can be applied to any loss.

## 1. Introduction

Dimensionality reduction is an ubiquitous and important form of data analysis. Recovering the inherent manifold structure of data—i.e. the local directions of large versus minimal variation—enables useful representations based on encoding highly varying directions. Not only can this reveal important structure in data, and hence support visualization, it also provides an automated form of noise removal and data normalization that can aide subsequent data analysis.

Although *linear* dimensionality reduction is a well studied topic, recent progress continues to be made with the investigation of convex regularizers, such as the trace norm or 2,1-norm, that enable application to general losses beyond squared error (Candes et al., 2011; Xu et al., 2010; Srebro & Shraibman, 2005). The literature on *nonlinear* dimensionality reduction, by comparison, has grown more rapidly yet devoted relatively less attention to developing appropriate regularizers. This provides one of our main motivations.

The focus of this paper is on *unsupervised* dimensionality reduction; that is, we will not directly address supervised variants, e.g. (Weinberger & Saul, 2009). We will also focus on *non-parametric* formulations that do not require an explicit map connecting the low and high dimensional representations; e.g. as in restricted Boltzmann machines (Larochelle & Bengio, 2008), auto-encoders (Rifai et al., 2011), or parameterized kernel reductions (Wang et al., 2010).

The primary benefit of focusing on unsupervised, non-parametric formulations is that it allows a simple yet comprehensive overview of current methods. In particular, we observe that nearly all current non-parametric methods can be expressed as regularized loss minimization of a reconstruction matrix followed by singular value truncation. This perspective allows us to distinguish the role of the *loss* from that of the *regularizer*: the loss relates the learned reconstruction to the data, whereas the regularizer relates the reconstruction to the desired topology *independent* of the data. Such a separation allows a simpler organization of current methods than current overviews (Burges, 2010; Lee & Verleysen, 2010a). More importantly, it reveals new research directions. A brief overview of current losses, for example, reveals a useful alternative that remains uninvestigated. Similarly, an assessment of current regularizers reveals that very few have been explored: in fact, only one family of convex regularizers has been widely used in the nonlinear case (distance maximization), which has known weaknesses. Although non-convex regularizers have been proposed to mitigate these weaknesses, these introduce intractability. Our main contribution is to derive efficient new convex regularizers that are able to combine distance maximization with rank reduction.

## 2. Preliminaries

Below we will need to manipulate and relate *data*, *kernel*, and *Euclidean distance* matrices respectively. Assume one is given $t$ observations, either expressed as a $t \times n$ data matrix $X$; a $t \times t$ kernel matrix $K$ where $K = K'$ and $K \succcurlyeq 0$; or a $t \times t$ squared Euclidean distance matrix $D$ where $D = D'$, $D \geq 0$, $\boldsymbol{\delta}(D) = \mathbf{0}$ and $HDH \preccurlyeq 0$, such that $\boldsymbol{\delta}$ denotes *diagonal* and $H = I - \frac{1}{t}\mathbf{1}\mathbf{1}'$ denotes the centering matrix. Then we can map between these various matrices via

$$\mathcal{K}(X) = XX' \tag{1}$$
$$\mathcal{D}(X) = \boldsymbol{\delta}(XX')\mathbf{1}' + \mathbf{1}\boldsymbol{\delta}(XX')' - 2XX' \tag{2}$$
$$\mathcal{D}(K) = \boldsymbol{\delta}(K)\mathbf{1}' + \mathbf{1}\boldsymbol{\delta}(K)' - 2K \tag{3}$$
$$\mathcal{K}(D) = -\frac{1}{2}HDH \tag{4}$$

(where the function intended is determined by the argument). Note that it is easy to map a data matrix to its corresponding kernel or Euclidean distance matrix, but such a map is neither linear nor invertible: kernel matrices drop orientation, while Euclidean distance matrices drop orientation and translation. However, by *centering* the data or kernel matrix, thus removing translation information, the relationship between kernel and Euclidean distance matrices becomes simple.

**Proposition 1** *A linear bijection exists between centered kernel and squared Euclidean distance matrices.*

It is easy to verify that $\mathcal{D}(\mathcal{K}(D)) = D$ and $\mathcal{K}(\mathcal{D}(K)) = HKH$ for a valid Euclidean distance and kernel matrix respectively. Therefore if $f(D)$ is convex in $D$ then $f(\mathcal{D}(K))$ must be convex in $K$. Proposition 1 thus allows one to equivalently re-express problems in terms of centered kernel matrices or Euclidean distance matrices without affecting expressiveness or convexity.

In this paper we assume the target dimensionality $d$ is fixed beforehand. That is, we are not addressing the problem of estimating the intrinsic dimensionality; for a survey see (Lee & Verleysen, 2010a, Ch.3). We will also need to make use of the indicator function

$$[\![\phi]\!] = \begin{cases} 0 & \text{if the predicate } \phi \text{ is true} \\ \infty & \text{if the predicate } \phi \text{ is false} \end{cases} . \tag{5}$$

## 3. Background: Linear Case

First briefly consider *linear* dimensionality reduction, which illustrates some basic points. Here one is given a *data matrix* $X$ and seeks a reduced rank representation $\hat{X}$. It turns out that a simple, generic strategy covers almost all methods that have been proposed: First, solve the regularized loss minimization problem

$$\min_{\tilde{X}} L(\tilde{X}; X) + R(\tilde{X}) \tag{6}$$

for a given loss $L$ and regularizer $R$, obtaining the reconstruction $\tilde{X}$. Then recover the low rank representation $\hat{X}$ by truncating all but the top $d$ singular values of $\tilde{X}$; that is, $\hat{X} = \tilde{U}_{:,1:d}\tilde{\Sigma}_{1:d,1:d}\tilde{V}'_{:,1:d}$ such that $\tilde{U}\tilde{\Sigma}\tilde{V}' = \tilde{X}$ is the singular value decomposition. Note that the loss relates $\tilde{X}$ to the data, $X$, whereas the regularizer enforces assumptions on the reconstruction $\tilde{X}$ that are independent of the data. Interestingly, it is the *regularizer*, not the loss, that typically determines the computational difficulty of this problem.

**Regularizers:** The role of the regularizer is to encourage the desired topology. To illustrate, consider the common regularizers proposed for the linear case

$$R_1(\tilde{X}) = [\![\text{rank}(\tilde{X}) \leq d]\!] \tag{7}$$
$$R_2(\tilde{X}) = \alpha\|\tilde{X}\|_{tr} \tag{8}$$
$$R_3(\tilde{X}) = \alpha\|\tilde{X}'\|_{2,1} \tag{9}$$

where $\alpha \geq 0$ is a regularization parameter. All of these clearly encode a desire for reduced rank.

In particular, the rank indicator (7) is the standard regularizer for spectral dimensionality reduction, which eliminates the need for truncation. Unfortunately rank is not convex, and enforcing (7) is only known to be tractable for squared loss (10) (Jolliffe, 1986). For other losses, such as absolute error (11) or Bregman divergence (12), rank is normally enforced by means of alternating descent in a factored representation: $\min_{AB} L(AB; X)$ where $A$ and $B$ are $t \times d$ and $d \times n$ respectively (Collins et al., 2001; Gordon, 2002). Unfortunately, this cannot guarantee optimality.[1]

The difficulty of working with rank explains the emergence of convex, rank-reducing regularizers such as the trace norm (8) (Candes et al., 2011; Srebro & Shraibman, 2005) and block norm (9) (Xu et al., 2010). In fact, the trace norm is known to be the tightest convex approximation to rank.[2] These regularizers allow a tractable formulation for general convex losses, and also allow a desired rank to be enforced by appropriately choosing $\alpha$ (Cai et al., 2008).

**Loss Functions:** Despite the importance of regularization, it is interesting to observe that until recently much of the research effort in linear dimensionality reduction investigated alternative losses, including

$$L_1(\tilde{X}; X) = \|X - \tilde{X}\|_F^2 \tag{10}$$
$$L_2(\tilde{X}; X) = \|X - \tilde{X}\|_{1,1} \tag{11}$$
$$L_3(\tilde{X}; X) = B_F(\tilde{X}\|X), \text{ such that} \tag{12}$$
$$B_F(\tilde{X}\|X) = F(\tilde{X}) - F(X) - \text{tr}(\nabla F(X)'(\tilde{X} - X)). \tag{13}$$

Here $\|\cdot\|_F$ denotes Frobenius norm, $\|\cdot\|_{1,1}$ denotes 1,1 block norm, and (13) is a Bregman divergence associated with strictly convex potential $F$. Beyond the squared error (10) used for PCA (Jolliffe, 1986), the absolute loss (11) has been recently proposed for robust PCA (Candes et al., 2011; Xu et al., 2010), and Bregman divergences (12) have been implicitly proposed in exponential family PCA (Collins et al., 2001; Gordon, 2002).[3] Interestingly, all these standard loss functions are *convex* in $\tilde{X}$.

---

[1] Random projection has also been considered, but it is difficult to establish guarantees for general losses (Brinkman & Charikar, 2003).

[2] Specifically, it is the bi-conjugate of the rank function over the unit sphere in spectral-norm (Recht et al., 2010).

The conclusion we draw from the linear case that the loss functions considered are standard, convex, and not the source of computational difficulty. Instead, it is *regularization* that has posed the greatest difficulty, particularly the desire to reduce rank. Interestingly, for the *non-parametric* case we find the same holds: almost every method follows regularized loss minimization plus truncation, almost every loss adopted is standard (if expressed between Euclidean distance matrices), and the main difficulty lies in devising regularizers that encourage desired topology. Such a simple perspective is surprisingly not widely appreciated.

## 4. Nonlinear Case

A general non-parametric approach to dimensionality reduction can be obtained by expressing the problem in terms of kernel matrices. (Recall that data or Euclidean distance matrices can always be converted to kernel matrices.) Here we assume one is given a $t \times t$ kernel matrix $K$ determined by the data and seeks a reconstruction $\tilde{K}$ that allows a reduced rank representation. Here too it turns out that a simple, generic strategy covers almost all methods that have been proposed: First, solve the regularized loss minimization

$$\min_{\tilde{K}=\tilde{K}',\tilde{K}\succcurlyeq 0,\tilde{K}\mathbf{1}=\mathbf{1}} L(\mathcal{D}(\tilde{K}); \mathcal{D}(K)) + R(\tilde{K}) \quad (14)$$

for a given loss $L$ and regularizer $R$, obtaining the reconstruction $\tilde{K}$. Then recover the low rank representation $\hat{X}$ by truncating all but the top $d$ eigenvalues and factoring; that is, $\hat{X} = \tilde{Q}_{:,1:d} \tilde{\Lambda}_{1:d,1:d}^{1/2}$, where $\tilde{Q}\tilde{\Lambda}\tilde{Q}' = \tilde{K}$ is the eigenvalue decomposition of $\tilde{K}$.

(The reason for stating the loss in (14) in terms of $\mathcal{D}(\tilde{K})$ will be explained below. Note that if the loss function $L$ is convex in its first argument it must also be convex in $\tilde{K}$ by Proposition 1.)

Although the difference between the linear and non-parametric formulations does not appear large, (14) offers far greater flexibility in recovering alternative topological structures, such as nonlinear manifolds in the original data. In fact, a key intuition behind most non-parametric dimensionality reduction methods is *unfolding*, where one supposes the data lay on a low di-mensional curved submanifold that is to be "unfolded" into a linear subspace. Unfortunately, current proposals conflate the role of the loss and regularizer in such a process, and propose only specific combinations of the two. We find it revealing to consider them separately.

---

[3] Noting the equivalence between regular exponential families and Bregman divergences (Banerjee et al., 2005).

### 4.1. Regularizers

The natural role for *regularization* in dimensionality reduction is to relate the reconstruction $\tilde{K}$ to a desired topology, expressing prior assumptions about the nature of the target representation independent of the data. For example, one might seek a coordinate representation in a linear subspace, in which case it is desirable to encourage "flattening" by spreading distances. However, regularization can express other topologies (see below). As before, the computational challenges appear to be primarily dictated by the regularizer.

The most common target topology is a linear subspace, for which the most common regularizers considered are

$$\begin{aligned} R_1(\tilde{K}) &= [\![\mathrm{rank}(\tilde{K}) \le d]\!] & (15) \\ R_2(\tilde{K}) &= \beta \mathrm{tr}(\tilde{K}) & (16) \\ R_3(\tilde{K}) &= -\alpha \mathrm{tr}(\tilde{K}) & (17) \end{aligned}$$

where $\alpha \ge 0$ and $\beta \ge 0$ are regularization parameters.

For example, the rank indicator (15) is the most commonly used regularizer, often associated with classical spectral dimensionality reduction (kernel PCA) using squared error (22) (Schoelkopf et al., 1999; Ham et al., 2004). Unfortunately, rank is not convex, and efficient training procedures are not known for other losses. Instead, for other losses, rank is typically enforced by resorting to local minimization in a factored representation: $\min_{\hat{X}} L(\mathcal{D}(\hat{X}\hat{X}'); D)$ where $\hat{X}$ is a $t \times d$ representation matrix. Unfortunately, no current loss provides a convex formulation in $\hat{X}$, and optimal solutions usually cannot be guaranteed.

Consequently, *convex* regularizers have also played a prominent role in non-parametric dimensionality reduction. For example, applying the trace norm here yields (16). Unfortunately, $\mathrm{tr}(\tilde{K}) = \mathbf{1}'\mathcal{D}(\tilde{K})\mathbf{1}/(2t)$ for centered $\tilde{K}$, thus minimizing trace is equivalent to *shrinking* distances; in opposition to the desire to unfold manifolds. Consequently, the negated regularizer (17) has proved more effective, forming one of the key components of maximum variance unfolding (MVU) (28) (Weinberger et al., 2004; 2007). Neither of these regularizers is completely satisfactory however.

***Partitioned regularizer:*** It is obvious what one would desire: to spread distances in the top $d$ dimensions and shrink distances in the remaining dimensions, for the target dimensionality $d$. Such a regu-larizer has been proposed by (Shaw & Jebara, 2007):

$$R_4(\tilde{K}) = -\alpha \max_{P \in \mathcal{P}} \text{tr}(P\tilde{K}) + \beta \min_{P \in \mathcal{P}} \text{tr}((I-P)\tilde{K}) \quad (18)$$

$$= \min_{P \in \mathcal{P}} \beta \text{tr}(\tilde{K}) - (\alpha+\beta)\text{tr}(P\tilde{K}), \text{ where} \quad (19)$$

$$\mathcal{P} = \{P : P' = P, I \succcurlyeq P \succcurlyeq 0, \text{tr}(P) = d\}. \quad (20)$$

Unfortunately, (19) is not convex; (Shaw & Jebara, 2007) resort to alternating minimization between $\tilde{K}$ and $P$. *Below we provide new convex regularizers that approximate (19), providing our main contribution.*

**Topographic Methods:** As an aside, it is interesting to note that alternative topologies can be encouraged via regularization. For example, given a graph expressed as an adjacency matrix $G \in \{0,1\}^{t \times t}$, a regularizer can encourage $\tilde{K}$ to adopt $G$'s structure

$$R_5(\tilde{K}) = -\max_{M \in \{0,1\}^{t \times t}, M\mathbf{1}=\mathbf{1}, \mathbf{1}'M=\mathbf{1}'} \text{tr}(M'\tilde{K}MG), \quad (21)$$

which provides a generalized approach to topographic embedding (Quadrianto et al., 2010; Bishop et al., 1998). Unfortunately, (21) is *concave* in $\tilde{K}$ and the inner optimization over $M$ is an NP-hard quadratic assignment problem, so we do not pursue this further.

### 4.2. Loss Functions

The role of the loss function is to relate the reconstruction to the *data*. In the formulation (14) we have chosen to express losses as between *Euclidean distance matrices*. We will see that such a viewpoint, although nonstandard, provides clarity. For example, unfolding is naturally enabled by loss *locality*: errors in reconstructing small target distances should be punished more harshly than errors in larger distances. Loss locality allows larger distances in the reconstruction more leeway to adapt to the desired target topology.

Expressing current losses in terms of distances reveals that they are almost all standard, convex, and obvious. We briefly survey standard losses to demonstrate how broadly the perspective applies to current methods, and to highlight useful alternatives.

**Classical Losses:** The oldest losses used for dimensionality reduction do not express locality, and therefore tend to recover linear subspace representations:

$$L_1(\hat{D}; D) = \|H(D - \hat{D})H\|_F^2 \quad (22)$$
$$L_2(\hat{D}; D) = \|D - \hat{D}\|_F^2 \quad (23)$$
$$L_3(\hat{D}; D) = \|\sqrt{D} - \sqrt{\hat{D}}\|_F^2 \quad (24)$$
$$L_4(\hat{D}; D) = \|D - \hat{D}\|_{1,1} \quad (25)$$

Here $\sqrt{\cdot}$ is applied component-wise. These losses are all convex in $\hat{D}$ (Dattorro, 2012, Ch.7). The doubly centered squared loss (22) is used in kernel PCA (Schoelkopf et al., 1999) (recall $HDH = -2\mathcal{K}(D)$). Although (23) is frequently mentioned in the multidimensional scaling (MDS) literature (Cox & Cox, 2001), its use is rare since it cannot be tractably combined with rank (15) (Dattorro, 2012). The absolute loss (25) in an important alternative that has been used in robust MDS (Cayton & Dasgupta, 2006).

**Local Losses:** Unlike the classical losses, however, local losses encourage unfolding by de-emphasizing errors on large distances. Let $\mathcal{N}(D) \in \{0,1\}^{t \times t}$ denote an adjacency function, such that $\mathcal{N}(D)_{ij} = 1$ indicates that $i$ and $j$ are neighbors in $D$ (i.e. either within a distance of $\epsilon$ or among the $k$ nearest neighbors). The best known examples of local losses are

$$L_5(\hat{D}; D) = \sum_{ij}(D_{ij} - \hat{D}_{ij})^2/D_{ij} \quad (26)$$
$$L_6(\hat{D}; D) = \sum_{ij}(D_{ij} - \hat{D}_{ij})^2 \, w(\hat{D}_{ij}) \quad (27)$$
$$L_7(\hat{D}; D) = \sum_{ij} [\![\mathcal{N}(D)_{ij} = 1 \text{ and } \hat{D}_{ij} \neq D_{ij}]\!] \quad (28)$$
$$L_8(\hat{D}; D) = \sum_{ij} \mathcal{N}(D)_{ij}(D_{ij} - \hat{D}_{ij})^2. \quad (29)$$

These are the Sammon loss (26) (Sun et al., 2011; Lee & Verleysen, 2010a); the curvilinear components loss (27) (Sun et al., 2010); the neighborhood indicator used in MVU and Isomap (28) (Weinberger et al., 2004; Tenenbaum et al., 2000); and the relaxed loss introduced in regularized MVU (29) (Weinberger et al., 2007), respectively. All such losses emphasize errors on small target distances (or predictions) over errors on large target distances. Moreover, they are all convex in $\hat{D}$.[4] Although the convexity of these losses with respect to $\hat{D}$ has not always received significant notice in the literature, this is an important fact for (14).

**Bregman Divergences:** Bregman divergences provide another loss specification that emphasizes locality. Recall that a Bregman divergence is defined by $B_F(\hat{D}\|D) = F(\hat{D}) - F(D) - \text{tr}(\nabla F(D)'(\hat{D} - D))$ for a strictly convex differentiable potential $F$, which by construction must be convex in $\hat{D}$. A number of such divergences have proved to be important in the dimensionality reduction literature, including

$$B_{F_9}(\hat{D}_{ij}\|D_{ij}) = \hat{D}_{ij}(\log \hat{D}_{ij} - \log D_{ij}) + D_{ij} - \hat{D}_{ij} \quad (30)$$
$$B_{F_{10}}(\hat{D}_{ij}\|D_{ij}) = \exp(-\hat{D}_{ij}/\sigma) - \exp(-D_{ij}/\sigma)$$
$$\qquad\qquad + \exp(-D_{ij}/\sigma)(\hat{D}_{ij} - D_{ij})/\sigma \quad (31)$$
$$B_{F_{11}}(\hat{D}_{i:}\|D_{i:}) = p(D_{i:})(\log p(D_{i:}) - \log p(\hat{D}_{i:}))' \quad (32)$$
$$\text{such that } p(D_{i:}) = \frac{\exp(-D_{i:})}{\exp(-D_{i:}\mathbf{1})} \quad (33)$$

---

[4] The convexity of curvilinear components loss (27) depends on $w(\hat{d})$; for example $\exp(-\hat{d}/\sigma)$, $1_{(\hat{d} \leq \epsilon)}$, or $1/\hat{d}$ (Lee & Verleysen, 2010a). The latter makes (27) convex.

$$B_{F_{12}}(\hat{D}\|D) = \text{tr}(p(D)'(\log p(D) - \log p(\hat{D}))) \quad (34)$$

$$\text{such that } p(D) = \frac{\exp(-D)}{\mathbf{1}'\exp(-D)\mathbf{1}}, \quad (35)$$

where $\sigma > 0$ is a scale parameter, log and exp are applied component-wise, and it is assumed the divergences are summed over all $ij$ or all $i$ where necessary.

The unnormalized entropy (30) was proposed in (Sun et al., 2011) to approximate the Sammon loss (26), whereas the reciprocal exponential Bregman divergence (31) was proposed in (Sun et al., 2010) to approximate the curvilinear components loss (27) under $w(\hat{d}) = \exp(\hat{d}/\sigma)$. However, the latter approximation was achieve by placing $\hat{D}$ in the *second* position, using $\ell(\hat{D}_{ij}; D_{ij}) = B_F(D_{ij}\|\hat{D}_{ij})$, which is no longer convex (see below). The Bregman divergence (32) matches the loss used in SNE (van der Maaten & Hinton, 2008) up to a minor variation: the transfer $p$ in SNE does not normalize over the entire vector, but only over the vector minus the current entry. The later, symmetric SNE error can also be recovered (almost) from the matrix-wise Bregman divergence (34) up to the same minor variation, plus a second exception: even though $p(\hat{D})$ is computed as in (35), $p(D)$ is computed by averaging the column and row probabilities through $ij$ using (33).

*Surprisingly, the exponential divergence (31) has not previously been used with $\hat{D}$ in the first, convex position. This yields a highly localized loss that is well suited to manifold unfolding, demonstrating even stronger locality than the other divergences. Therefore, we investigate its behavior further below.*

**Large Margin Losses:** Large margin losses for nonlinear dimensionality reduction have also been proposed in (Shaw & Jebara, 2009):

$$L_{13}(\hat{D};D) = \sum_i \max_j \bar{\mathcal{N}}(D)_{ij}(\max_k \mathcal{N}(D)_{ik}\hat{D}_{ik} - \hat{D}_{ij})$$

$$\text{where } \bar{\mathcal{N}}(D)_{ij} = 1 - \mathcal{N}(D)_{ij} \quad (36)$$

$$L_{14}(\hat{D};D) = \max_{N \in \mathfrak{N}} \ell(N, \mathcal{N}(D)) + \text{tr}(\hat{D}(\mathcal{N}(D) - N)) \quad (37)$$

where $\ell$ is a local margin loss function. The intuition behind the loss (36) is simple: for each node $i$, one would like the distances to all disconnected nodes $j$ (such that $\mathcal{N}(D)_{ij} = 0$) to be greater than the distance to the furthest connected node $k$ (i.e. such that $\mathcal{N}(D)_{ik} = 1$). The second loss (37) is defined with respect to a class of alternative adjacency matrices $\mathfrak{N}$ producible by running an efficient dynamic program on candidate distance matrices $\hat{D}$. This loss, termed "structure preserving" in (Shaw & Jebara, 2009), behaves like a structured output SVM loss that tries to make sure that the sum of estimated distances on $\mathcal{N}(D)$ are less than on any alternative adjacency matrix in $\mathfrak{N}$, plus a margin dictated by how far $N \in \mathfrak{N}$ is from $\mathcal{N}(D)$. Both losses are convex.

**Non-convex Losses:** Finally, even though non-convex losses are computationally problematic, a few important methods are expressible in this manner:

$$L_{15}(\hat{D}; D) = \sum_{ij} p(D)_{ij}(\log p(D)_{ij} - \log q(\hat{D})_{ij}) \quad (38)$$

$$\text{where } q(\hat{D})_{ij} = \frac{(1+\hat{D}_{ij})^{-1}}{\sum_{k \neq l}(1+\hat{D}_{kl})^{-1}} \quad (39)$$

$$L_{16}(\hat{D}; D) = -\max_{W: \boldsymbol{\delta}(W) = \mathbf{0}, W_{\bar{N}} = 0, W\mathbf{1} = \mathbf{1}}$$
$$\text{tr}\Big((I-W)((1-\rho)D + \rho\hat{D})(I-W)'\Big) \quad (40)$$

where $\rho \geq 0$ is a weighting parameter. The first loss corresponds to tSNE (38), which is a modification of the symmetric SNE loss (34), using the same transfer $p(D)$ on the target distances $D$, but using a "heavier tailed" transfer function on $\hat{D}$ (van der Maaten & Hinton, 2008). The second loss (40) can be shown to be equivalent to local linear embedding (LLE) (Roweis & Saul, 2000; Saul & Roweis, 2003; Ham et al., 2004) if one tracks the solution in the limit as $\rho \searrow 0$. Clearly, this formulation shows that the LLE loss is highly non-convex in $\hat{D}$.

Note that other non-parametric dimensionality reduction methods can also be expressed in terms of regularized minimization of a loss between distance matrices, but the above suffices to illustrate how comprehensively current methods can be covered. Interestingly, this loss-based framework generalizes probabilistic formulations (Lawrence, 2011), since e.g. large margin losses cannot be naturally expressed as log-likelihoods.

## 5. New Convex Regularizers

Our main contribution is to propose two new convex regularizers for non-parametric dimensionality reduction. In particular, we formulate convex relaxations of the partitioned regularizer (19), which simultaneously seeks to spread distances on the top $d$ dimensions while shrinking distances in the remaining directions. The consequences for both rank reduction and manifold unfolding are clear. We first introduce a slight modification of (19) by adding a small quadratic smoother

$$R_5(\tilde{K}) = \min_{P \in \mathcal{P}} \frac{\gamma}{2}\text{tr}(\tilde{K}^2) + \beta\text{tr}(\tilde{K}) - (\alpha+\beta)\text{tr}(P\tilde{K}), \quad (41)$$

where $\gamma > 0$, and $\mathcal{P}$ is the same as defined previously in (20). Although this modified regularizer is still not convex in $\tilde{K}$, it enables two useful relaxations.

### 5.1. Completed Square

The first relaxation we propose is extremely simple: (41) can be made jointly convex in $\tilde{K}$ and $P$ simply by completing the square, yielding

$$R_6(\tilde{K}) = \min_{P \in \mathcal{P}} \beta \mathrm{tr}(\tilde{K}) + \frac{\gamma}{2} \left\| \tilde{K} - \frac{\alpha+\beta}{\gamma} P \right\|_F^2. \quad (42)$$

This is guaranteed to be an upper bound on (41) since we are merely adding a nonnegative term $\frac{\gamma}{2}\|\frac{\alpha+\beta}{\gamma}P\|_F^2$. A lower bound on (41) can be recovered by subtracting $\frac{d(\alpha+\beta)^2}{2\gamma}$ from (42) since $\|P\|_F^2 \leq d$ for all $P \in \mathcal{P}$.

The main benefit of this relaxation is that it is extremely simple and computationally attractive: a simple modification of the alternating minimization strategy of (Shaw & Jebara, 2007) now yields a global solution. This does not, however, yield the tightest convex approximation of (41), as we now demonstrate.

### 5.2. Bi-conjugation

Recall that the conjugate of a function $f$ is defined by $f^*(y) = \sup_{x \in \mathrm{dom}(f)} \langle x, y \rangle - f(x)$ (Borwein & Lewis, 2006). Importantly, any function is lower bounded by its bi-conjugate; i.e. $f \geq f^{**}$ (Borwein & Lewis, 2006, §4.2). Therefore, a general strategy for deriving maximal convex lower bounds on objective functions can be based on *Fenchel bi-conjugation* (Jojic et al., 2011). Here, we obtain a new regularizer by formulating the tightest convex relaxation of (41) based on its bi-conjugate and define

$$R_7(\tilde{K}) = R_5^{**}(\tilde{K}). \quad (43)$$

By construction, this must be a convex function in $\tilde{K}$.

**Theorem 1**

$$R_5^{**}(U) = \max_{Z=Z'} \mathrm{tr}(UZ)$$
$$- \frac{1}{2\gamma} \max_{P \in \mathcal{P}} \|[Z - \beta I + (\alpha+\beta)P]_+\|_F^2, \quad (44)$$

where $[\cdot]_+ = \max(0, \cdot)$ is applied to the eigenvalues.

*Proof:* We compute the Fenchel bi-conjugate of

$$g(K) = \begin{cases} R_5(K) & \text{if } K \succcurlyeq 0 \\ \infty & \text{otherwise} \end{cases}. \quad (45)$$

First, the conjugate of $g(K)$ is easily derived as

$$g^*(Z) = \max_{K \succcurlyeq 0} \mathrm{tr}(Z'K) - R_5(K) \quad (46)$$
$$= \max_{K \succcurlyeq 0} \max_{P \in \mathcal{P}} \mathrm{tr}(Z'K) - \frac{\gamma}{2}\mathrm{tr}(K^2) - \beta\mathrm{tr}(K)$$
$$+ (\alpha+\beta)\mathrm{tr}(KP). \quad (47)$$

Note that the domain of $g^*$ is actually all $t \times t$ matrices, but (47) implies that $g^*(Z) = g^*(Z') = g^*((Z+Z')/2)$, hence for the purpose of computing $g^{**}$ no generality is lost if we restrict the domain of $g^*$ to symmetric matrices. Now for any $U \succcurlyeq 0$ we obtain

$$g^{**}(U) = \max_{Z \in \mathcal{S}^t} \mathrm{tr}(UZ) - g^*(Z) \quad (48)$$
$$= \max_{Z \in \mathcal{S}^t} \min_{P \in \mathcal{P}} \min_{K \succcurlyeq 0} \mathrm{tr}(UZ) - \mathrm{tr}(ZK)$$
$$+ \frac{\gamma}{2}\mathrm{tr}(K^2) + \beta\mathrm{tr}(K) - (\alpha+\beta)\mathrm{tr}(KP)$$
$$= \max_{Z \in \mathcal{S}^t} \min_{P \in \mathcal{P}} \mathrm{tr}(UZ) -$$
$$\max_{K \succcurlyeq 0} \mathrm{tr}[K(Z - \beta I + (\alpha+\beta)P)] - \frac{\gamma}{2}\|K\|_F^2 \quad (49)$$
$$= \max_{Z \in \mathcal{S}^t} \mathrm{tr}(UZ) - \frac{1}{2\gamma} \max_{P \in \mathcal{P}} \|[Z-\beta I+(\alpha+\beta)P]_+\|_F^2.$$

The last equality is due to von Neumann's trace inequality (Borwein & Lewis, 2006) and the elementary equality $\max_{x \geq 0} xy - \frac{\gamma}{2}x^2 = \frac{1}{2\gamma}(y)_+^2$. ∎

Putting $\gamma = 0$, Theorem 1 implies that the Fenchel biconjugate of (19) is exactly (17) (for $Z = -\alpha I$) if $d > 0$ and is (16) (for $Z = \beta I$) if $d = 0$, which might explain the success of MVU (Weinberger et al., 2004).

Although (44) appears to be a complex regularizer, it is not computationally much harder to optimize than (42). To evaluate $R_5^{**}(U)$, an optimal $Z$ in (44) must be solved, but this is a convex problem and admits efficient optimization. First note that the inner maximization in (44) necessarily attains its maximum at some extreme point of the set $\mathcal{P}$ (for we are maximizing a convex function). Next inspecting (49) and invoking von Neumann's trace inequality one more time we reduce (44) to its vector counterpart. Let $\{z_i\}$ and $\{u_i\}$ (both arranged in decreasing order) be the eigenvalues of $Z$ and $U$, respectively, then we just need to solve:

$$\min_{z_i \geq z_{i+1}} -(2\gamma)\mathbf{u}'\mathbf{z} + \sum_{i=1}^{d}[z_i + \alpha]_+^2 + \sum_{i=d+1}^{t}[z_i - \beta]_+^2. \quad (50)$$

Temporarily ignoring the constraints, the optimal $\mathbf{z}$ is obvious since all elements of $\mathbf{z}$ are separated in (50):

$$\hat{z}_i = \begin{cases} \gamma u_i - \alpha & \text{if } i \leq d \\ \gamma u_i + \beta & \text{if } i \geq d+1 \end{cases}. \quad (51)$$

Now observe that $\hat{z}_i$ automatically satisfies the order constraints in the two blocks, thus the only possibility of violating the order constraint is between the blocks, i.e. $\hat{z}_d < \hat{z}_{d+1}$. But if we knew $\hat{z}_d = \lambda$, we would be able to fix the order by setting

$$z_i(\lambda) = \begin{cases} \max(\hat{z}_i, \lambda) & \text{if } i \leq d \\ \min(\hat{z}_i, \lambda) & \text{if } i \geq d+1 \end{cases}. \quad (52)$$

| Data set | Loss | tr($K$) | $-$tr($K$) | partition | partition($\gamma$) | bi-conjugate | compl. sq. |
|---|---|---|---|---|---|---|---|
| Swiss | rMVU | 246M (33) | 1.00 (0.2) | 1.00 (5) | 1.00 [0.0] (5) | 0.33 [-1.7] (6) | 0.95 [15] (6) |
| | SNE | 7.82 (42) | 1.00 (14) | 1.69 (4) | 1.69 [1K] (4) | 0.61 [143] (100) | 0.64 [442] (74) |
| | expBreg | 0.00 (1) | 1.00 (1) | 0.47 (5) | 0.47 [1.3] (5) | 0.00 [-1.7] (6) | 0.00 [15] (7) |
| Face | rMVU | 36K (15) | 1.00 (0.2) | 1.00 (3) | 1.00 [0.01] (3) | 0.64 [-0.6] (47) | 0.82 [17] (2) |
| | SNE | 1.31 (31) | 1.00 (12) | 1.69 (6) | 1.69 [974] (6) | 1.03 [486] (35) | 0.83 [503] (32) |
| | expBreg | 0.27 (0.6) | 1.00 (0.3) | 1.00 (3) | 1.00 [0.03] (2) | 0.27 [-0.1] (1) | 0.34 [17] (2) |
| 3d-sin | rMVU | 112K (25) | 1.00 (0.1) | 1.00 (6) | 1.00 [0.01] (4) | 0.25 [-2.0] (6) | 0.80 [19] (7) |
| | SNE | 1.22 (35) | 1.00 (4) | 2.00 (2) | 2.00 [757] (2) | 0.76 [116] (29) | 0.70 [296] (23) |
| | expBreg | 0.39 (1) | 1.00 (0.5) | 1.01 (2) | 1.01 [2.4] (2) | 0.41 [-1.6] (3) | 0.42 [20] (2) |
| Brush | rMVU | 252 (2) | 1.00 (0.3) | 1.61 (0.4) | 1.61 [0.3] (0.4) | 0.41 [-1.2] (1) | 0.42 [59] (0.3) |
| | SNE | 0.99 (2) | 1.00 (0.4) | 2.43 (0.7) | 2.43 [38] (0.6) | 0.98 [-14] (7) | 0.79 [70] (2) |
| | expBreg | 0.06 (0.2) | 1.00 (0.1) | 0.55 (0.6) | 0.55 [1.1] (0.6) | 0.07 [-1.3] (1) | 0.07 [59] (0.4) |
| mean | | 204M | 1.00 | 1.29 | 1.29 | 0.48 | 0.57 |

Table 1. *Relative* reconstruction losses using different losses (SNE (34), regularized MVU loss (29), exponential divergence (31)) and six different regularizers (from left to right: (16), (17), (19), (41), (44), (42)). The trace maximizing regularizer (17) is being used as the reference. In each case $d = 2$, $\alpha = 1$, $\beta = 0.1$, $\gamma = 10^{-3}$, and losses were scaled by 10. In the three leftmost columns, the number ($s$) in parentheses gives the run time in seconds. In the three rightmost columns, the numbers $[o](s)$ in the parentheses give the objective function value $[o]$ and run time in seconds ($s$) respectively.

Therefore we only need to find $\lambda$ that minimizes (50), which is a univariate convex piecewise quadratic function in $\lambda$, hence can be done very quickly.

To summarize, for evaluating $R_5^{**}(U)$ (and obtaining a subgradient), we need to perform SVD on $U$ and then solve (50). The former costs $O(t^3)$ while the latter costs $O(t)$, where $t$ is the number of points.

## 6. Experimental Evaluation

Most evaluations of dimensionality reduction methods resort to subjective assessments of specific case studies. However, many proposals exist for quantitatively evaluating the quality of different methods in a somewhat objective manner (Sun et al., 2011; Lee & Verleysen, 2010b; van der Maaten, 2009). Evaluation is clearer when the regularizer and loss components are considered separately. In particular, to compare *regularizers*, an objective assessment can be based on the loss values achieved by the low-rank reconstruction. *For a given loss function $L$*, we measure the reconstruction loss $L(\mathcal{D}(\hat{X}\hat{X}'); D)$ achieved by the recovered low rank representation $\hat{X}$. Another objective assessment can be based on the run time of the corresponding methods.[5] Finally, we can assess the quality of the convex relaxations by measuring the gap between the final objective values achieved using the relaxed regularizers versus their source regularizer $R_5$ (41).

[5] Here we measure run time using a common implementation based on accelerated projected subgradient descent (Tseng, 2008). Accelerated projected subgradient descent proves to be a far more scalable and generally applicable method for these problems than a generic SDP solver.

In our experiments, we compare different regularizers on a representative set of loss functions. The losses we considered are: regularized MVU loss (29), SNE (34), and the reciprocal exponential Bregman divergence (31). The results are given in Table 1 for six regularizers and three losses on four different data sets.[6] The regularizers used are $R_2$ (16), $R_3$ (17), $R_4$ (19), $R_5$ (41), $R_6$ (42), and $R_7$ (44). Additionally, we report the objective values for $R_5$ and its relaxations $R_6$ and $R_7$, to assess the approximation gap.

In terms of reconstruction error, the new convex regularizers obtain the best reconstruction errors overall, almost always outperforming the other competitors. A particular disadvantage of the partitioned regularizers is that they are not convex, hence sensitive to initialization: the objective values (shown in the square brackets in the rightmost three columns) demonstrate that inferior local minima do exist, since the completed square regularizer ($R_6$) provides an upper bound on the optimal partition($\gamma$) objective. The run times for the new regularizers are somewhat slower.

## 7. Conclusion

We have presented simple view of non-parametric dimensionality reduction by separating the roles of the regularizer and loss function. The result is a compact survey of a large fraction of the literature, which reveals a natural loss function that has not been thoroughly explored. More importantly, we developed two new convex relaxations of a useful, but non-convex reg-ularizer. We investigated the behavior of the new regularizers across a representative sample of loss functions. Important directions for future research include extending the convex regularization framework to consider other target topologies, sparsity, and mixtures.

[6] Data set details are given in the supplement.